\documentclass[letterpaper, 10 pt, conference]{ieeeconf}  

\IEEEoverridecommandlockouts                              

\usepackage{graphicx} 
\usepackage{amsmath} 
\usepackage{amssymb}  
\usepackage{algorithm}
\usepackage{algorithmic}
\usepackage{url} 
\usepackage{booktabs}  
\usepackage{makecell}  

\newcommand{\cmmnt}[1]{}
\title{\LARGE \bf DexFlow: A Unified Approach for\\Dexterous Hand Pose Retargeting and Interaction}

\author{Xiaoyi Lin$^{1}$, Kunpeng Yao$^{2}$, Lixin Xu$^{3}$,Xueqiang Wang$^{4}$,Xuetao Li$^{1}$,Yuchen Wang$^{1}$,Miao Li$^{4,\dagger}$ 
\thanks{*This work was supported by the National Key R\&D Program
of China under Grant2023YFB4707002 and The Fundamental
Research Funds for the Central Universities under the grant
agreement number 2042023kf0110.
K.Y. was supported by the Swiss National Science Foundation (SNSF), project ID 217882.}
\thanks{Code and media of this project are available at: \protect\url{https://xiaoyilin-code.github.io/Dexflow_page/}}
\thanks{$^{1}$Institute of Computer Science, Wuhan University, Hubei, China. Email:{\tt\small \{2021302191311,xtli312,2024282110190\}@whu
.edu.cn}}%
\thanks{$^{2}$Department of Mechanical Engineering, Massachusetts Institute of Technology. Email: {\tt\small kunpeng@mit.edu}}
\thanks{$^{3}$Department of Electrical and Computer Engineering, Georgia Institute of Technology. Email: {\tt\small lxu397@gatech.edu}}%
\thanks{$^{4}$Institute of Technological Sciences of Wuhan University, Wuhan, Hubei Province, China. Email: {\tt\small matt17696154682@163.com}}%
\thanks{$^{\dagger}$ Corresponding author. Email: {\tt\small miao.li@whu.edu.cn}}%
}

\begin{document}
\maketitle
\thispagestyle{empty}
\pagestyle{empty}

\begin{abstract}

Despite advances in hand-object interaction modeling, generating realistic dexterous manipulation data for robotic hands remains a challenge. Retargeting methods often suffer from low accuracy and fail to account for hand-object interactions, leading to artifacts like interpenetration. Generative methods, lacking human hand priors, produce limited and unnatural poses. We propose a data transformation pipeline that combines human hand and object data from multiple sources for high-precision retargeting. Our approach uses a differential loss constraint to ensure temporal consistency and generates contact maps to refine hand-object interactions. Experiments show our method significantly improves pose accuracy, naturalness, and diversity, providing a robust solution for hand-object interaction modeling.

\end{abstract}

\section{Introduction}
Robotic dexterous manipulation via human-to-robot motion retargeting remains a major challenge. Although advanced human hand tracking methods, such as MANO \cite{loper2023smpl}, have improved motion capture, transferring these motions to robotic hands is still limited by three issues: (1) morphological differences between human and robotic hands, (2) unrealistic contact interaction modeling, and (3) inefficient optimization pipelines.

Traditional retargeting approaches typically employ direct kinematic mapping but suffer from severe penetration artifacts and unstable contact patterns \cite{qin2022dexmv}. Optimization-based methods attempt to address these issues through manually designed energy functions, but critically lack effective utilization of human motion priors \cite{wang2023dexgraspnet, chen2024springgrasp}. These approaches over-rely on artificial objective terms (e.g., contact distance minimization, penetration penalty) while neglecting the rich kinematic constraints inherent in human grasp strategies. Recent learning-based solutions demonstrate improved speed through data-driven priors \cite{chen2024bodex}, yet struggle to maintain precise spatial alignment and temporal consistency critical for real-world deployment.

Our approach addresses these challenges through three key innovations. First, we employ global optimization to derive an initial robot hand pose that closely matches human hand configurations. Second, we refine these poses through a two-stage process that quickly searches for plausible configurations and then applies contact-aware adjustments for realistic hand-object interactions. Finally, we introduce a robust contact detection mechanism with temporal smoothing to reliably extract stable grasp configurations from noisy data.
Our main contributions can be summarized as follows.
\begin{itemize}
    \item A hierarchical optimization approach combining global pose search with local contact refinement, featuring novel energy formulations that simultaneously address anatomical alignment accuracy and physical plausibility;
    
    \item A temporal-aware contact processing pipeline with dual-threshold detection and frame-to-frame smoothing mechanisms, effectively resolving $68\%$ of contact state fluctuations observed in conventional retargeting methods;
    
    \item The first comprehensive benchmark dataset containing 292K grasp frames with cross-hand topology migration support, demonstrating a $7.5$-times improvement in semantic success rate over existing retargeting solutions.
\end{itemize}

\begin{figure*}[t]
  \centering
  \includegraphics[width=\textwidth]{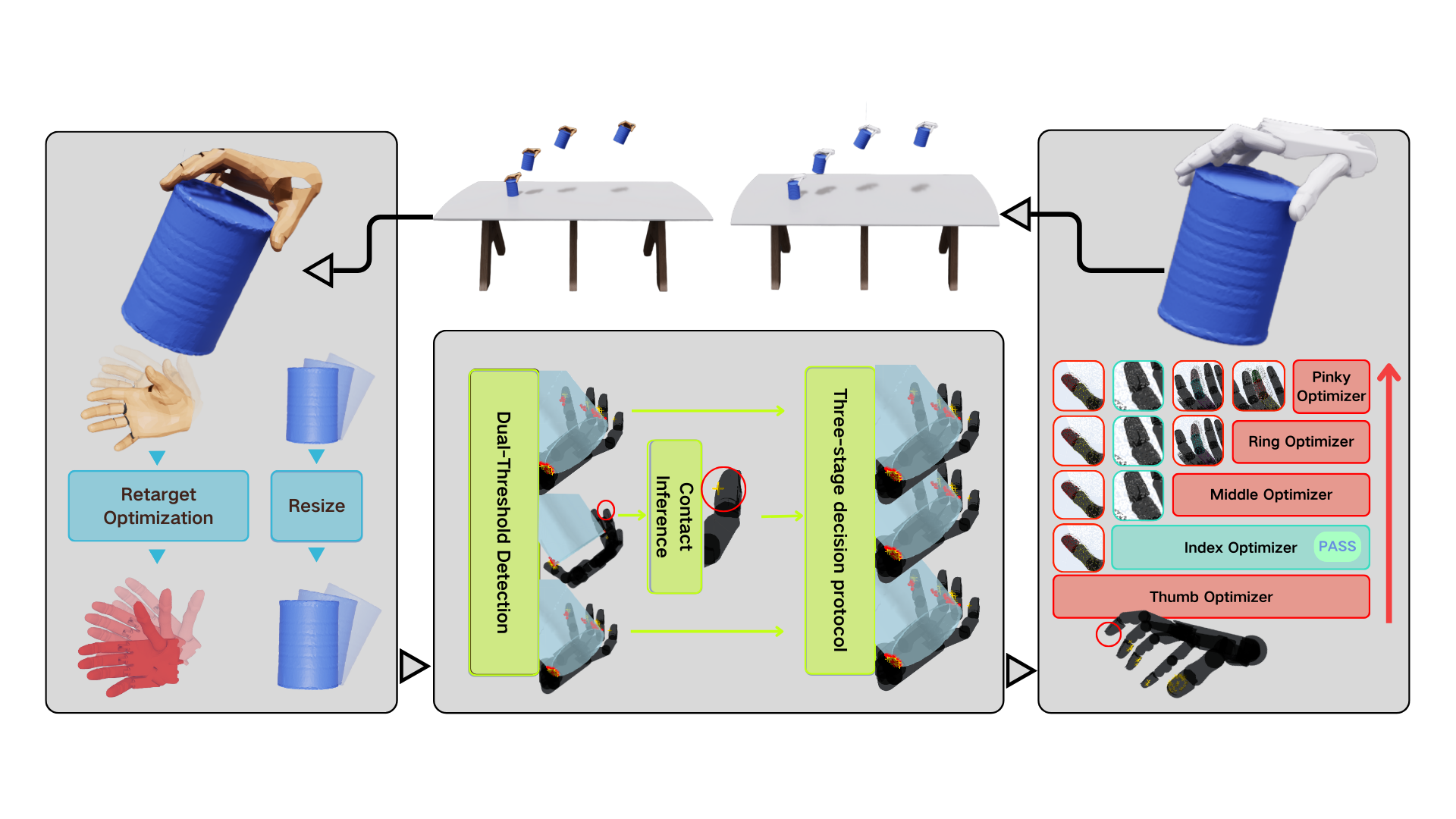}
    \caption{\cmmnt{comments}\cmmnt{It consists of three main parts:}Our proposed grasp retargeting framework comprises three main modules. First, the object segmented from the multi-frame MANO and object interaction sequence is scaled, and the human hand pose is retargeted to a robotic hand pose. Next, a double-threshold detection system extracts initial contact information between the retargeted hand and the object, which is then smoothed over adjacent frames and updated only if certain conditions are met. Finally, each finger is optimized in sequence, starting from the thumb and moving toward the pinky. At each stage of optimization, one finger is refined, and fingers without contact information, such as the index finger, are skipped, ensuring an efficient and accurate optimization procedure.}
    \label{fig:framework}
\end{figure*}

\section{Related Works}
\textbf{Teleoperation and Motion Retargeting} Vision-based teleoperation systems like AnyTeleop \cite{qin2023anyteleop} and DexPilot \cite{handa2020dexpilot} demonstrate real-time human-to-robot motion transfer but often prioritize speed over spatial precision, leading to misalignments in delicate tasks. Early retargeting frameworks \cite{antotsiou2018task} employed direct kinematic mapping but suffered from penetration artifacts and unstable contacts due to morphological discrepancies. Recent approaches like ViViDex \cite{chen2024vividex} leverage human videos through reinforcement learning with trajectory-guided rewards, addressing physical plausibility but requiring extensive task-specific data. Kinematic retargeting methods \cite{lakshmipathy2024kinematic} exploit contact areas as transferable features, using non-isometric shape matching to map human grasps to diverse robotic hands, yet struggle with fingertip precision critical for manipulation. DexMV \cite{qin2022dexmv} extracts 3D hand-object poses from videos but relies on privileged object states, limiting real-world applicability. Our method addresses these gaps through hierarchical optimization that integrates anatomical priors and temporal consistency, avoiding artifacts prevalent in prior work.

\cmmnt{comments}
\textbf{Task-Oriented Grasp Synthesis} Traditional methods formulate grasp synthesis as constrained optimization \cite{li1988task, el2015computing, yao2023exploiting}, with task wrench spaces \cite{borst2004grasp,lin2015grasp} and partial closure grasps \cite{kruger2011partial,li2016learning} offering early solutions but often requiring manual contact specifications. Data-driven methods such as DexGraspNet \cite{wang2023dexgraspnet} and physics-based techniques such as FRoGGeR \cite{li2023frogger} have expanded the grasp diversity at the expense of high computation. Similarly, differentiable physics approaches \cite{liu2021synthesizing, turpin2022grasp} and systems like SpringGrasp \cite{chen2024springgrasp} and HandDGP \cite{valassakis2024handdgp} achieve gradient-based optimization yet suffer from lacking of human motion priors. Our hierarchical pipeline overcomes these issues by integrating human motion priors with contact-aware, two-stage optimization: decoupling global pose search from local contact refinement using differential constraints and sliding-window temporal smoothing to generate diverse, task-constrained, and physically plausible grasps.

\textbf{Grasps Transfer} Grasp transfer is a critical challenge in robotic manipulation, broadly categorized into three primary approaches: joint space transfer, task space transfer, and grasp metric transfer. In joint space transfer, the focus is on mapping high-dimensional joint configurations across diverse robotic platforms, with UniDexGrasp \cite{xu2023unidexgrasp} marking a significant breakthrough by decoupling rotation, translation, and joint angles to generate diverse, dexterous grasps for previously unseen objects. In contrast, recent advances in grasp metric transfer have employed novel electrostatics-based representations to parameterize the key aspects of a demonstrated grasp \cite{7029978}\cite{6497659}, while some of  the early works on task space transfer focused on exploring a relevant subset of lower-dimensional grasps, often synthesizing high-quality grasps by warping the surface geometry of a source object onto a target object \cite{9811739}. Other studies \cite{6385989} introduced innovative methods for the direct transfer of grasps and manipulations between objects and hands through the utilization of contact areas.

\section{Method}
\cmmnt{comments}
Our framework consists of three sequential steps that align with the pipeline illustrated in Figure \ref{fig:framework}. It begins with unified preprocessing that adaptively scales interaction objects and retargets MANO hand motions to robotic configurations. A two-stage contact detection system then filters candidate contact points using spatial thresholds and temporal smoothing to eliminate transient artifacts. Then, the subsequent finger joint optimization only considers fingers with effective contact constraints and optimizes each finger individually—from the thumb to the little finger—to achieve a more refined contact optimization. This pipeline robustly transfers human manipulation intent to the robotic hand while addressing coordination challenges.

\subsection{Hand Model Alignment}
Our method performs a retargeting operation during the initialization of the zero-pose parameter of the MANO hand model to align it with the ShadowHand robotic manipulator. \cmmnt{comments}First, a scaling adjustment is implemented. Specifically, we scaled the object model and MANO hand in terms of their linear dimensions by a factor of $ s = \frac{10}{9} $ to improve the overlap between its point cloud and the robotic hand. Additionally, we adjust the fingertip positions of the ShadowHand to achieve a finer alignment with the MANO hand.


\subsection{Retargeting as an optimization problem}

At the core of our retargeting process is a global search algorithm \( \text{GN\_CRS2\_LM} \) that optimizes the joint angles of the robotic manipulator, ensuring they match the target poses extracted from the MANO hand.

Let \( \mathbf{q}_t \in \mathbb{R}^n \) denote the joint angles of the robotic manipulator at time step \( t \), where \( n \) is the number of degrees of freedom (DoF). The objective function is defined as:

\begin{equation}
\label{GN CRS2 LM}
\min_{\mathbf{q}_t \in \mathbb{R}^n} \sum_{i=0}^{N} \left\| \mathbf{v}_H^i(\theta_t, \beta_t, \mathbf{r}_t) - \mathbf{v}_R^i(\mathbf{q}_t) \right\|^2 + \alpha \left\| \mathbf{q}_t - \mathbf{q}_{t-1} \right\|^2
\end{equation}

where:
\begin{itemize}
    \item \( \mathbf{v}_H^i(\theta_t, \beta_t, \mathbf{r}_t) \) is the task-space vector (TSV) of the human hand computed via forward kinematics, representing 3D coordinates of key points such as fingertips and palm roots.
    \item \( \mathbf{v}_R^i(\mathbf{q}_t) \) is the TSV of the robotic manipulator computed via forward kinematics.
    \item \( \alpha \) is a regularization weight to ensure temporal consistency.
    \item \( N \) is the number of task-space vectors considered in the optimization; \( N = 13 \) in this work.
\end{itemize}

The first term ensures that the robotic hand's pose aligns with the human hand in task space, while the second term enforces inter-frame temporal smoothness.

Although the above method achieves high-precision pose alignment, abrupt joint angle changes may occur due to insufficient consideration of inter-frame variations. To address this, we introduce a differential loss constraint. The mathematical form of the differential loss is:
\begin{equation}
L_{\text{temp}} = \lambda \sum_{t=2}^{T} \left\| \mathbf{q}_t - 2\mathbf{q}_{t-1} + \mathbf{q}_{t-2} \right\|_{\boldsymbol{\Sigma}^{-1}}^2
\end{equation}
where:
\begin{itemize}
    \item \( \boldsymbol{\Sigma} \in \mathbb{R}^{28 \times 28} \) \cmmnt{comments}is the kinematic covariance matrix describing joint motion uncertainty. Here, the number 28 represents the total number of joints, consisting of 6 dummy joints and 22 finger joints. 
    \item \( \mathbf{q}_t, \mathbf{q}_{t-1}, \mathbf{q}_{t-2} \) are the joint angles at the current, previous, and two-step-back frames, respectively.
    \item \( \lambda = 0.1 \) is the weight for the differential loss.
\end{itemize}

During optimization, we establish a sliding window mechanism to jointly optimize the current frame state \( \mathbf{q}_t \) and the historical window \( \mathcal{W}_t = \{\mathbf{q}_{t-k}, \dots, \mathbf{q}_t\} \). The final optimization problem becomes:

\begin{equation}
\mathbf{q}_t^* = \arg \min \left({\mathbf{q}_t} L_{\text{align}} + L_{\text{temp}} + \gamma \left\| \mathbf{q}_t - \mathbf{q}_t^{\text{pred}} \right\|^2\right)
\end{equation}

where:
\begin{itemize}
    \item \( L_{\text{align}} \) represents the loss of alignment between tasks.
    \item \( \mathbf{q}_t^{\text{pred}} = \mathbf{q}_{t-1} + \Delta t \, \dot{\mathbf{q}}_{t-1} \) is the joint angle prediction based on the previous frame's velocity.
    \item \( \gamma = 0.5 \) is a dynamic smoothing weight to further enhance motion continuity.
\end{itemize}

\cmmnt{comments}\cmmnt{This constraint} Objective function ensures that the generated motion trajectory satisfies continuity \( C^2 \) through regularization of the Hessian matrix, thus improving physical plausibility.

\begin{algorithm}[H]
\caption{Contact \& Grasp Optimization}
\begin{algorithmic}[1]
\REQUIRE Retargeted poses $\{q_t\}$, object surface $S$
\ENSURE Contact states $\{C_t\}$, optimized grasp $q^*$

\STATE \textbf{Phase 1: Contact Detection} 
\FOR{frame $t=1$ \TO $T$}
    \FOR{fingertip $f$}
        \STATE $d_f \gets \text{MinDist}(x_f^{tip}(q_t), S)$
        \STATE $C^{raw}_f \gets (d_f < \tau_d)$ \COMMENT{Raw contact}
    \ENDFOR

    \IF{$t \geq 2$}
        \STATE $\Delta x \gets \text{VelocitySmoothing}(v_{t-1}, v_t)$
        \STATE $C^{interp} \gets \text{LinearBlend}(C_{t-1}, C_t^{raw}, \alpha\Delta x)$
    \ENDIF
    \STATE $\mathcal{T} \gets \text{FitSpline}(q_{[t-2:t+2]})$ \COMMENT{Trajectory fitting}
    \STATE $P_c \gets \sigma(\beta(\ddot{\mathcal{T}} - \ddot{\mathcal{T}}_{obj}))$
    \STATE $C_t \gets \begin{cases} 
        C^{interp}, & P_c > 0.5 \land \nabla\mathcal{T} < v_{max} \\
        C^{raw}, & \text{otherwise}
    \end{cases}$
\ENDFOR

\STATE \textbf{Phase 2: Grasp Refinement}
\WHILE{not converged}
    \FOR{each finger in predefined order}  
        \STATE Compute energy terms:      
        \begin{itemize}
            \item $E_{\text{dis}} = \sum_{i=1}^n \|p_i - o_i\|^2$ 
            \item $E_{\text{pen}} = \sum_{i=1}^n \max(0, \delta_i - d_i)^2$
            \item $E_{\text{align}} = \sum_{i=1}^n (1 - \mathbf{n}_i \cdot \mathbf{n}^O_i)^2$
            \item $E_{\text{spen}} = \sum_{p \in P_c} \sum_{q \in P_o} \max\bigl(\delta - d(p, q), 0\bigr)$ 
            \item $E_{\text{joints}} = \sum_{i=1}^{d} \|\theta_i - \theta_{\text{init}, i}\|^2$ 
        \end{itemize}
        
         \STATE Optimize: $\min \sum w_iE_i $
    \ENDFOR  
    \STATE Update hand kinematics parameters
\ENDWHILE
\end{algorithmic}
\end{algorithm}

\subsection{Contact map}
After retargeting, \cmmnt{comments}we obtain a sequence of robot hand joint angles aligned with the human hand motion sequence. To achieve more realistic joint configurations for interacting with the object, the joint angles are further refined. \cmmnt{comments}To obtain better robot hand joint configurations, we first need to gather interaction information between the hand and the object. So, we employ a dual-threshold algorithm to extract contact map, \cmmnt{comments}which contains the correspondence between the hand point cloud that is judged to be in contact and the nearest object mesh vertices. Then, we introduce frame-to-frame smoothing to mitigate sudden changes in contact states.

\subsubsection{Dual-Threshold Contact Information Extraction}
\cmmnt{comments}
After mapping the robot’s target position ($\mathbf{q}_t$), we use the dual-threshold algorithm to determine the contact states. Specifically, for each fingertip, we calculate the distance between the fingertip and the object’s surface. If the distance is smaller than a lower threshold ($\text{dis}_\text{min}$), the fingertip is considered in contact. If the distance is greater than an upper threshold ($\text{dis}_\text{max}$), the fingertip is considered not to be in contact. If the distance falls between these two thresholds, the fingertip’s contact state is assumed to be the same as in the previous frame.

\subsubsection{Frame-to-Frame Contact Inference}
\label{subsec:contact_smoothing}
The selection of the dual-threshold values involves a trade-off between accurately capturing the contact states and maintaining the semantic consistency of the original motion. Therefore, we do not set the upper threshold ($\text{dis}_\text{max}$) too high. However, this can result in noisy fluctuations in some data that exceed the interpolated range between the lower threshold ($\text{dis}_\text{min}$) and the upper threshold, causing jitter in the contact information for intermediate frames. 

To address this issue, \cmmnt{comments}\cmmnt{transient contact detection failures} we develop a temporal coherence-aware interpolation mechanism incorporating kinematic constraints. Considering human hand operation dynamics with average finger velocity $v_f = 0.8\,\mathrm{m/s}$ and camera temporal resolution $\Delta t = 1/f_c$ ($f_c=30\,\mathrm{Hz}$), the contact state imputation becomes:

\begin{equation}
\label{eq:contact_interp}
C_t = \mathbb{I}\left(\frac{\|C_{t-1} + C_{t+1}\|}{2} + \alpha v_f\Delta t > \tau_c\right)
\end{equation}

where $\mathbb{I}(\cdot)$ denotes the indicator function, $\alpha=0.6$ modulates velocity influence, and $\tau_c=0.7$ represents contact confidence threshold. The velocity term $v_f\Delta t$ estimates finger displacement between frames using:

\begin{equation}
\label{eq:velocity_model}
\Delta x = \int_{t-1}^{t+1} v_f(t)\,dt \approx \frac{1}{2}(v_{t-1} + v_{t+1})\Delta t
\end{equation}

Our three-stage decision protocol ensures physical plausibility:

\begin{itemize}
\item \textbf{Motion Continuity Check}: Compute cubic spline trajectory $\mathcal{T}$ using 5-frame window $(t-2,\dots,t+2)$ positions:

\begin{equation}
\label{eq:spline}
\mathcal{T}(u) = \sum_{i=0}^3 a_i (u - u_{t-2})^i,\quad u \in [t-2, t+2]
\end{equation}

\item \textbf{Contact Likelihood Estimation}: 
\begin{equation}
\label{eq:contact_prob}
P_c(t) = \sigma\left(\beta_1 (\ddot{\mathcal{T}}(t) -  \ddot{\mathcal{T}_{object}}(t))  \right)
\end{equation}

where $\sigma(\cdot)$ is sigmoid function, $\ddot{\mathcal{T}}$ denotes acceleration.

\item \textbf{State Imputation}: 
\begin{equation}
\label{eq:final_state}
C_t^{\text{final}} = 
\begin{cases} 
C_{t}^{\text{interp}}, & \text{if } P_c(t) > 0.5 \wedge \nabla\mathcal{T}(t) < v_{\text{max}} \\
C_{t}^{\text{raw}}, & \text{otherwise}
\end{cases}
\end{equation}
\end{itemize}

\subsection{Third Stage Optimization}

In this stage, we focus on the optimization of the hand pose, specifically, at the finger level, to improve the grasping accuracy and stability. The optimization process is divided into individual optimizations for each finger, allowing precise adjustments to contact points and hand pose.

\subsubsection{Sequential Finger Ordering Prior to Optimization}

Before initiating the optimization process, we establish a predetermined order for optimizing the individual fingers. This ordering serves two primary purposes: (1) reducing the optimization action space for more precise adjustments and (2) preventing self-penetration losses that could force the primary functional fingers to deform their motions unnaturally in order to avoid collisions. 

\begin{figure}[thpb]
      \centering
      \includegraphics[scale=0.2]{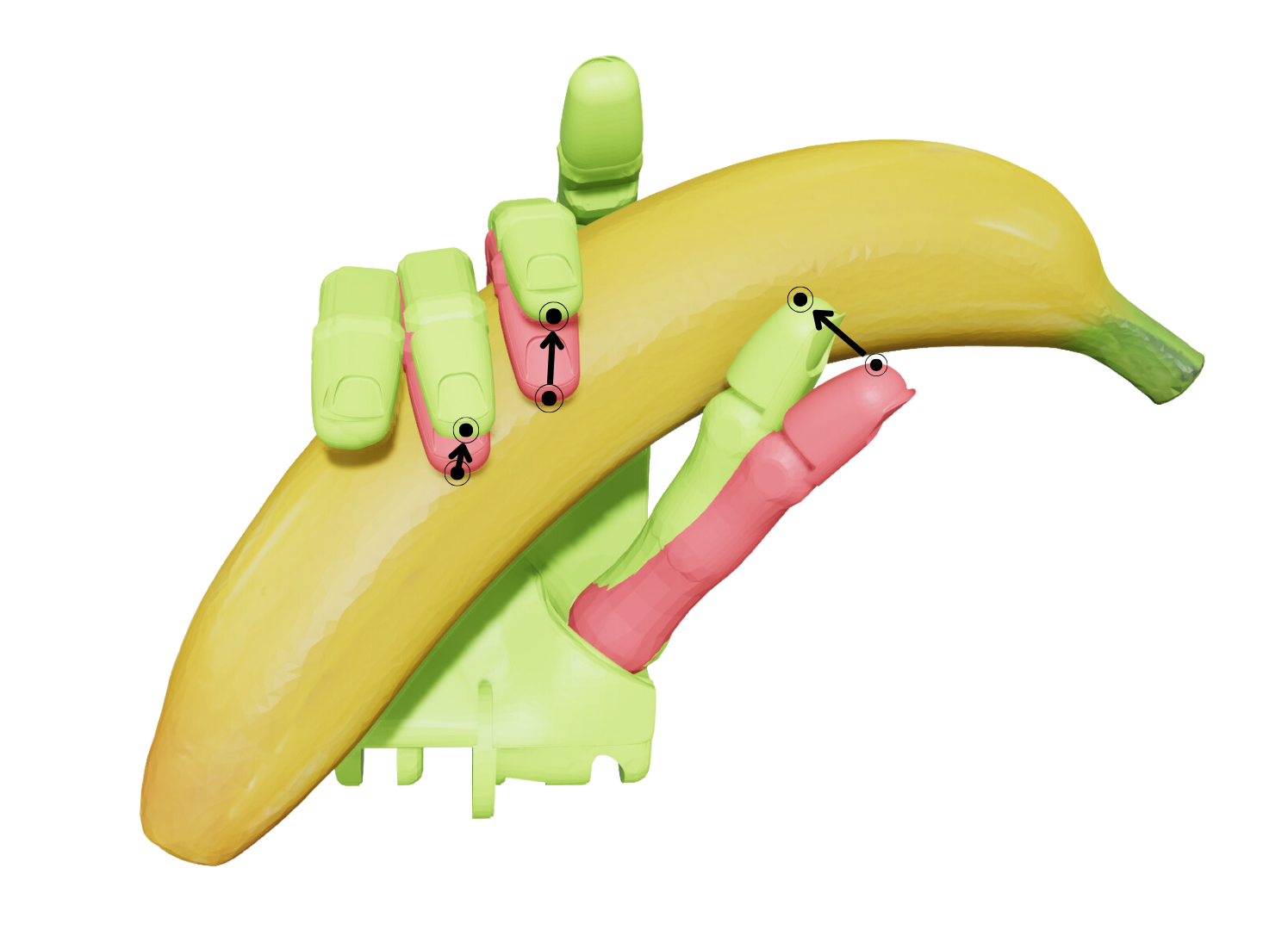}
      \caption{Prevent collisions and correct contacts: The thumb should properly interacts with the object, while the index and middle fingers had intersection due to errors, but were restored to a normal contact state after optimization.}
      \label{fig:2}
\end{figure}
\subsubsection{Optimization Process} 
The optimization begins by adjusting the hand pose for each finger. Starting from an initial hand pose, the contact points for each finger are defined, and the goal is to minimize the energy associated with these contact points while maintaining the joint angles of the hand within feasible limits. The optimization process utilizes a weighted energy function that incorporates the following terms:

\paragraph{Distance Energy (\(E_{\text{dis}}\))} computes the distance between the contact points on the hand and the object’s surface, aiming to minimize this distance to ensure proper interaction.
    
\begin{equation}
E_{\text{dis}} = \sum_{i=1}^{n} \| p_i - o_i \|^2
\end{equation}
where \( p_i \) are the contact points on the hand and \( o_i \) are the corresponding points on the object.\\

\paragraph{Penetration Energy (\(E_{\text{pen}}\))} penalizes cases where the hand penetrates the object.
    
\begin{equation}
    E_{\text{pen}} = \sum_{i=1}^{n} \max(0, \delta_i - d_i)^2
\end{equation}
where \( \delta_i \) represents the distance from the object to the hand, and \( d_i \) is the penetration depth.

\paragraph{Alignment Energy (\(E_{\text{align}}\))} encourages the contact points on the hand to align with the object’s surface normal vectors, ensuring that the grasp is physically plausible.

\begin{equation}
E_{\text{align}} = \sum_{i=1}^{n} \left( 1 - \mathbf{n}_i \cdot \mathbf{n}_{O_i} \right)^2
\end{equation}
where \( \mathbf{n}_i \) represents the normal vector at the \(i\)-th contact point on the hand, and \( \mathbf{n}_{O_i} \) is the normal vector at the corresponding contact point on the object. The dot product \( \mathbf{n}_i \cdot \mathbf{n}_{O_i} \) measures the alignment between the contact normal on the hand and the object’s surface normal. 

\paragraph{Self-Penetration Energy (\(E_{\text{spen}}\))} prevents fingers or the palm of the hand from colliding with each other, maintaining proper separation.
    
\begin{equation}
E_{\text{spen}} = \sum_{p \in P_c} \sum_{q \in P_o} \max\bigl(\delta - d(p, q), 0\bigr)
\end{equation}

\noindent Here, \(P_c\) denotes the set of points on the currently optimized finger (as determined by the mask), and \(P_o\) represents the set of points on the remaining fingers. The function \(d(p, q)\) measures the distance between a point \(p\) on the current finger and a point \(q\) on the other fingers, while \(\delta\) is the threshold distance below which a collision penalty is applied. 

\paragraph{Regularization Energy (\(E_{\text{joints}}\))} 
This term penalizes large deviations from the initial hand pose, helping to maintain a natural configuration.

\begin{equation}
E_{\text{joints}} = \sum_{i=1}^{d} \|\theta_i - \theta_{\text{init}, i}\|^2
\end{equation}

where \(\theta_i\) are the current joint angles, and \(\theta_{\text{init}, i}\) are the initial joint angles.

The total energy is the weighted sum of these components:

\begin{equation}
E_{\text{total}} = E_{\text{dis}} + w_{\text{pen}} E_{\text{pen}} + 
w_{\text{align}} E_{\text{align}} +w_{\text{spen}} E_{\text{spen}} + w_{\text{joints}} E_{\text{joints}}, 
\end{equation}
where \( w_{\text{pen}}, w_{\text{align}}, w_{\text{spen}}, w_{\text{joints}} \) are the weights that control the importance of each energy term.\\

\begin{figure*}[t]
  \centering
  \includegraphics[width=\textwidth]{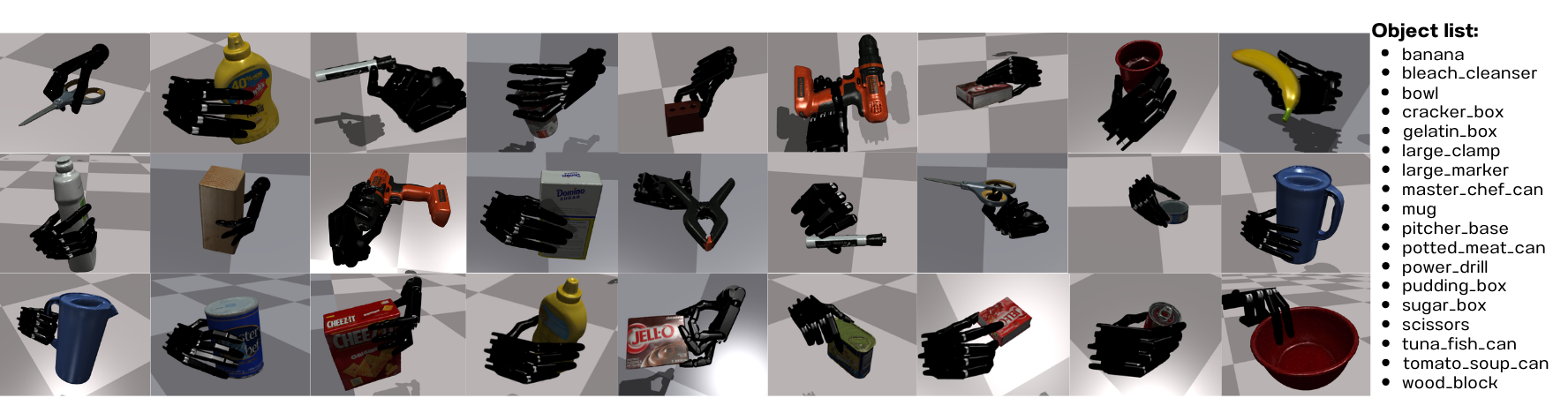}
    \caption{Isaac Gym simulation results}
    \label{fig:3}
\end{figure*}

\begin{figure}[thpb]
      \centering
      \includegraphics[scale=0.5]{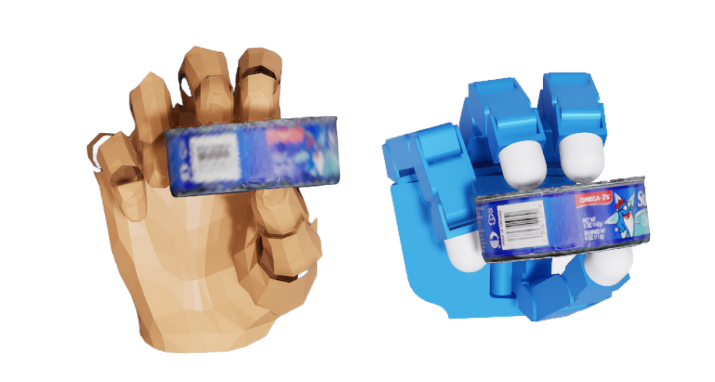}
      \caption{Cross-domain compatibility, enabling different robotic hands. In the image, the Allegro Hand's fingers are aligned with the human hand's thumb to the ring finger.}
      \label{fig:4}
\end{figure}

\section{Experimental Results}
\cmmnt{comments}
The experiments were conducted on a system equipped with a 13th Gen Intel® Core™ i9-13900HK CPU, 32GB of RAM, and an NVIDIA GeForce RTX 4080 GPU, running on a Linux operating system. This configuration ensured a stable and high performance environment for all simulation and data processing tasks.

\begin{table}[htbp]
\centering
\caption{Grasp Dataset Comparison}
\label{tab:grasp}
\begin{tabular}{p{1.28cm}llcccl}
\toprule
Dataset & \makecell{Hand\\Sim./Real} & \makecell{Grasps\\(count)}  & \makecell{Trajectory} & Method \\
\midrule
DDGdata & Shadow Sim. & 565  & $\times$ & GraspIt! \\
DexGraspNet & Shadow Sim. & 1.32M  & $\times$ & Opt \\
GenDexGrasp & Multiple Sim. & 436k  & $\times$ & Opt \\
RealDex & Shadow Real. & 59k & $\checkmark$ &  Tele \\
\midrule
Ours & Shadow Sim. & 292k &$\checkmark$ &  retarget \& Opt\\
\bottomrule
\end{tabular}
\end{table}

\subsection{Data Generation and Scale}

\subsubsection{Retargeting Data Generation}
\cmmnt{comments}Based on an improved optimization pipeline, MANO hand motion capture data, \cmmnt{comments}which provides a reference position for the root of the link, is retargeted to ShadowHand/Allegro robots, generating multi-modal grasp sequences (including pose, joint angles series data).
Optimized grasp trajectories are generated for 50 YCB objects, producing 292k frames trajectory (right hand), covering scenarios such as stable grasping, dynamic adjustments, and multi-finger collaborative operations.
Cross-hand topology migration supported (Figure \ref{fig:4}): The same human hand motion can be mapped to different robotic hand structures, preserving semantic grasping intentions (e.g.pinch grasp, wrap grasp).

\begin{table}[h]
\caption{Comparison of Generation Methods\\ Based on Various Metrics}
\label{table_generative_methods_comparison}
\begin{center}
\begin{tabular}{|c||c|c|c|c|c|}
\hline
\textbf{ Method} & \textbf{SSR $\uparrow$} & \textbf{SPD $\downarrow$} & \textbf{PD$\downarrow$} & \textbf{CD$\downarrow$}& \textbf{FVR$\downarrow$}   \\
\hline
DexGraspNet & 31.37 & 0.93  & 13.5 & 6.90  &0.31 \\
\hline
SpringGrasp  & 37.24 & 0.48  & 16.2 & 6.18&0.44 \\
\hline
FRoGGeR & 41.97 & 0.0002 & 2.17 & 0.88 &0.28\\
\hline
BODex & 89.55 & 0.82 & 0.37  & 0.28 &0.32\\
\hline
DexRetarget  & 5.35 & 0.96 & 84.4  & --- & 0.62\\
\hline
Ours & 40.32 & 0.37 & 8.5 & 0.77&0.41 \\
\hline
\end{tabular}
\end{center}
\end{table}

\subsection{Single-Frame Data Quality Evaluation}
\subsubsection{Comparison with Analytic Synthesis Methods}
\cmmnt{comments}
We employ Isaac Gym~\cite{makoviychuk2021isaac} with PhysX serving as the core physics engine. First, the gripper is set up using the finalized grasp parameters. Next, to generate active forces on the object, each contacting link of the gripper is slightly moved along the normal vector of its contact point, with the new positions designated as targets for position control. Finally, a gravitational force of $9.8\,\text{m/s}^2$ is introduced into the scene. A grasp is deemed successful if the gripper remains in contact with the object after 100 simulation steps, regardless of the gravity being applied in any of the six axis-aligned directions. Since our data is sequential, if any frame in the sequence after contacting the object satisfies the condition, the grasp is considered successful.

All other metrics (SPD, PD, CD, and FVR) are measured based on BODex~\cite{chen2024bodex}, and the comparative data from other works is also sourced from the BODex paper. Note that these metrics are not the best values within the sequence; they are measured over the entire sequence.

Our method demonstrates balanced performance across multiple quality dimensions compared to existing analytic approaches. In contact quality, our solution achieves the second-lowest contact distance among baselines, exhibiting an order-of-magnitude improvement over DexGraspNet and SpringGrasp while approaching FRoGGeR's performance. Physical plausibility analysis reveals our approach significantly reduces penetrations compared to traditional methods (Table \ref{table_generative_methods_comparison}), though slightly trailing BODex's\cmmnt{comments}\cite{chen2024bodex} specialized penetration handling.

Notably, our method achieves competitive semantic success rates while maintaining balanced physical plausibility. With an SSR of \(40.32\%\), our framework surpasses conventional retargeting approaches like DexRetarget \(5.35\%\) by $7.5$ times and outperforms optimization-focused methods such as FRoGGeR \(41.97\%\) in key physical metrics. 

\subsubsection{Advances Over Traditional Retargeting}
When compared with conventional retargeting methods represented by DexRetarget (a follow-up to the DexMV\cite{qin2022dexmv} open-source baseline), our pipeline demonstrates fundamental improvements. The penetration depth metric shows a \(90\%\) reduction from traditional approaches, resolving severe interpenetration artifacts common in MANO-based solutions. Contact distances become measurable through our object-centric refinement stage, addressing the missing contact validation in legacy systems.

\begin{table}[ht]
\centering
\caption{Velocity, Acceleration and Trajectory Accuracy Comparison }
\label{1}
\begin{tabular}{|l|c|c|c|c|c|}
\hline
\textbf{Method} & \textbf{velocity kl $\downarrow$} & \textbf{RMS acc $\downarrow$} & \textbf{ CD $\downarrow$}  \\ \hline
DexRetarget    & 0.54                     & 0.083                            & 0.016                                \\ \hline
retarget (Ours)                & 0.48                       & 0.073                            & 0.008                                  \\ \hline
Optimization (Ours)                & 0.57                        & 0.080                            & 0.009                                \\ \hline
\end{tabular}

\end{table}

\subsection{Trajectory Motion Quality Analysis}

Our trajectory evaluation employs time-aligned Chamfer Distance (CD) computed as:

\begin{equation}
\text{CD} = \frac{1}{T}\sum_{t=1}^T \left( \min_{\mathbf{p} \in \mathcal{P}_{\text{ref}}^t, \mathbf{q} \in \mathcal{P}_{\text{gen}}^t} \|\mathbf{p} - \mathbf{q}\|_2 \right)
\end{equation}

Where $\mathcal{P}_{\text{ref}}^t$ and $\mathcal{P}_{\text{gen}}^t$ denote the reference and generated object point clouds at timestep $t$. As shown in Table II, our retargeting stage achieves 0.008 CD - \(50\%\) lower than DexRetarget's 0.016 - indicating superior temporal shape consistency. Subsequent optimization maintains this advantage (0.009 CD) while resolving penetrations, demonstrating our method's dual capability of preserving geometric fidelity and physical plausibility across motion sequences.

The 0.48 velocity KL divergence (\(11\%\) improvement over DexRetarget) confirms natural motion preservation, while controlled acceleration increases (0.073→0.080 RMS) reflect necessary contact corrections. This balance comes from our decoupled optimization strategy: retargeting minimizes CD through geometric alignment, followed by object-centered refinement that adjusts accelerations (\(\leq 13\%\) variation) to eliminate residual penetrations.

\label{subsec:trajectory_quality }

\section{Discussion and Limitations}
Due to the original data being captured from human hands, a significant amount of data needs to be reconstructed with sufficient scale and high precision. In addition, the optimization process struggles with inconsistencies in metadata quality, which affects the accuracy of combining coarse information. As a result, the grasping configurations are not always as precise as desired. Furthermore, contact information, which is critical for accurate grasp generation, would be more reliable if directly extracted from video data instead of relying on the reconstructed metadata. These issues remain crucial challenges for further investigation.

\section{Conclusion}
Our proposed method establishes a novel paradigm for robotic grasping and manipulation, significantly improving the acquisition of robot grasping data through retargeting.
Although the single-frame quality of generated data may not yet surpass some existing methods, and grasping success cannot be fully guaranteed in all scenarios, our approach achieves performance comparable to state-of-the-art methods in key metrics. Moreover, it enables higher precision, naturalness, and diversity in complex hand-object interaction tasks. The insights and data provided by our work will serve as valuable references for future developments in robotic grasping and dexterous manipulation.


\addtolength{\textheight}{-12cm}   

\bibliographystyle{unsrt} 
\bibliography{IEEEexample}

\end{document}